# Evolutionary Algorithms for Fuzzy Cognitive Maps

Tsimenidis Stefanos

Abstract

Fuzzy Cognitive Maps (FCMs) is a complex systems modeling technique which, due to its unique advantages, has lately risen in popularity. They are based on graphs that represent the causal relationships among the parameters of the system to be modeled, and they stand out for their interpretability and flexibility. With the late popularity of FCMs, a plethora of research efforts have taken place to develop and optimize the model. One of the most important elements of FCMs is the learning algorithm they use, and their effectiveness is largely determined by it. The learning algorithms learn the node weights of an FCM, with the goal of converging towards the desired behavior. The present study reviews the genetic algorithms used for training FCMs, as well as gives a general overview of the FCM learning algorithms, putting evolutionary computing into the wider context.

## I. Introduction

Fuzzy Cognitive Maps (FCMs) were first proposed by Kosko [1], and are based on Axelrod's Cognitive Maps [2]. They're directed graphs that model complex systems and are used in decision-making, forecasting, and optimization. The nodes depict the system's concepts, and the directional edges identify the causal relationships among nodes. Each edge is represented by a direction arrow and a weight coefficient that quantifies, in the range [-1,1], the strength of the causal connection between two concepts. Zero weight signifies the absence of influences between the concepts.

FCMs accomplish the representation, in simplified form, of a system's structure, of the concepts that make it, and of the causal relationships among these concepts. Through the representation of the interrelated influences and the quantification of the influences among the nodes. FCMs can successfully capture the dynamic behavior of a system and predict the overall change effected to it by a change, or sequence of changes, in one or more concepts [3].

Systems of practical interest are usually complex and consist of interconnected subsystems and modules. Their behavior is non-linear and cannot be calculated by adding up the behaviors of their parts [4]. Real world systems are also characterized by fuzziness. Events take place not absolutely, but to a degree [5]. events contribute (as causes) to activate other events, not absolutely, but to a degree. It is this non-linearity and fuzziness of real world systems that FCMs were made to capture and model. They can, theoretically, model systems of arbitrary complexity, and their applications are almost endless.

They are widely used in economics [6], social science [7], environmental science [8] [9], bio-informatics [10], industrial control [11], medical science [12][13], optimization, etc. Also, their interpretability renders them useful for policy and decision making, as well as all aspects of pattern recognition since, in direct contrast with most traditional machine learning models, with FCMs we can know exactly how they reach a certain conclusion, and therefore we can evaluate them.

An FCM can be constructed either by getting expert opinion on a particular field, or by using data to automate the process. We can aggregate FCMs constructed either from expert knowledge or from various datasets into an integrated FCM. We can start from expert advice (either real persons or literature) and continue with knowledge extraction from data, combining the two approaches.

Training an FCM entails specifying the node weights, with the intention of either establishing the mapping between the FCM's inputs and outputs, getting the system to converge in a desired behavior. In regards to convergence, there are three possibilities: (1) the system converges to a stable state of equilibrium, (2) to an oscillation or periodic movement, or (3) a chaotic behavior around a chaotic attractor.

Learning algorithms for FCMs are divided into three categories. First, we have Hebbian-type learning, adopted from the traditional method of train artificial neural networks, and adapted for FCMs. Second, we have evolutionary algorithms that treat the training task as an optimization problem. Third, we have the hybrid approaches that combine Hebbian learning with evolutionary algorithms.

Real world systems are highly complex and dynamic. This renders FCMs, which were designed specifically for capturing complex dynamics, a powerful for modeling physical systems, but also poses a difficulty in their implementation. Identifying causal relationships may be a relatively simple and intuitive task, but specifying the degree of influence, ie the weight values of the edges, entails challenges. The more a system rises in complexity, the more costly the offsets of the weights of an FCM from the real-world system, and the less accurate the model. This is why research in FCM learning algorithms is of utmost importance and has risen in the last few years, with evolutionary algorithms holding, for reasons we shall see shortly, the top position.

The present study begins with the theoretical foundation of FCMs. Then follows an overview of the the learning algorithms that have been proposed and implemented, by category – Hebbian learning, evolutionary algorithms, and hybrid techniques. Then there is a discussion on evolutionary learning algorithms and, finally, conclusions.

## II. Fuzzy Cognitive Maps

FCMs are defined by the four following parameters:
C: the concepts/nodes of the system
W: two dimensional weight-matrix, one weight for each pair of concepts
A: two-dimensional matrix with the values of each concept (rows) for each discrete time interval (columns).
f: activation function, applied on A values and compresses them in a desired range.

When a weight value is positive, raising the value of the cause concept results in rise of the effected concept. When it is negative, raising the value of the cause concept results in lowering of the effected concept. When it is zero, the two concepts do not exert any influence on each other.

To compute the value A of a concept at some specific moment, we sum the values, from the previous moment, of all the concepts that influence the concept of interest, multiplied by the weights of the interconnections, and then we pass the result through the activation function.

This is the basic and most widely used method, but it is not the only one. Different approaches have been occasionally proposed, like a method [12] that takes under consideration the concept values of previous moments, and is used on concepts that aren't influenced by others, or a method [14] that alleviates saturation issues, where a concept value gets too close to the lower\upper limit of its accepted range. The choice of concept update method is made according to the requirements of each application.

FCMs can capture the feedback loops in complex systems, where a concept influences another and in turn is influenced by it. This gives a temporal dimension to the system, where the present state is dependent on the previous ones.

Another key characteristic of FCMs is the ease with which they can be finetuned and improved. New concepts can be added, relationships among concepts may change, different FCMs may be aggregated into an new integrated FCM. Systems of very high complexity can be divided into simpler subsystems and studied separately, modeled with separate FCMs and, finally, combined into a final, augmented FCM.

The activation function that squashes the node activation values into the range, depending on the application, either [0,1] or [-1,1], can be either discrete, or continuous. The most important activation functions are:

1. Bivalent function. It's discrete, with binary output, where $f(x)=0$ for $x>0$, and $f(x)=1$ for $x<=0$. Due to its binary output, the FCM will never yield chaotic behavior, but will either converge into equilibrium or exhibit periodic movement.
2. Trivalent function. Discrete, its output can have three possible values: -1 for $x<0$, 0 for $x=0$, and 1 for $x>0$. Similarly with the bivalent function, it may converge into either equilibrium or periodicity, but never into chaotic behavior.
3. Hyperbolic function. Continuous activation function, output values in the range [-1,1], and may lead to chaos.
4. Sigmoid function. Continuous function with acceptable values in the range

[0,1]. Takes parameters for the slope and the offset of the function.

The activation function is chosen according to the needs of the particular problem at hand. Fundamental limitation of discrete activation functions is their reduced capacity of complex system representation, as they can't express quantitative relationships. A disadvantage of continuous activation function is their possibility to lead to chaotic behavior. Continuous functions have been shown to yield better results [15], and among them, the sigmoid function exceeds the rest [16].

### III. FCM Training

The objective of FCM training is to specify the weight matrix of the system's edges. This can be done either through consulting with experts, or using historical data, or by combining the two approaches.

Three categories of approaches exist: Hebbian learning, evolutionary algorithms, and hybrid approaches that combine Hebbian learning with evolutionary computing.

With Hebbian learning we start from expert knowledge to construct the weight matrix so as to converge towards a desired state. With genetic algorithms human experts are replaced by historical data, and the weight matrix is inferred from them. With the hybrid techniques FCM learning is divided in two parts, first we generate a weight matrix through consulting with experts, and then we finetune it using historical data.

A. Hebbian Learning Algorithms

With this method, after we construct the initial weight matrix, we modify it with unsupervised learning. This approach is based on theoretical foundations set by Donald Hebb in his classic *The Organization of Behavior* [17], summed up in the famous expression "neurons that fire together, wire together."

The first implementation of Hebbian learning for FCMs, called Differential Hebbian Learning (DHL), was proposed by Dickerson and Kosko [5]. When the cause concept and the effected concept change values simultaneously, the weight that connects them is increased by a predetermined amount, and the connection between the two concepts is reinforced. Otherwise the weight will remain unchanged for the iteration. The weight values are updated iteratively.

One disadvantage of this method is that the weight between two nodes is modified without taking the system as a whole into consideration. Only local information is used, and the high-level information of the total system is wasted.

A proposal from Huerga [18], called Balanced Differential Algorithm (BDA), aims to improve learning by taking into account all the concepts that change simultaneously with the weight modification. Although the method brought better results, its application was limited.

Papageorgiou et al [19] developed the Active HL (AHL) algorithm, where experts determine the initial FCM structure and the weights get modified iteratively until a certain condition is fulfilled. This method can be used when the FCM's node values change in a specific order. In direct contrast with the previous algorithms, where only the non-zero weights get modified, now all the weights are influenced.

Another algorithm is the Nonlinear Hebbian Learning (NHL) [20]. the initial FCM structure, as introduced by the experts, remains unchanged during the entire process. The algorithm stops either when the system converges on equilibrium, or the desired state is approximated satisfactorily.

The previous two methods do not strive to modify the FCM to cover the input data entirely, but to make the model converge towards a desired state.

The Improved Nonlinear HL (INHL) algorithm that was proposed later [21] has an additional coefficient in the weight update rule, aiming to prevent the system from getting trapped in local minima.

Another proposal to improve NHL was developed by Stach et al and is called Data-Driven NHL (DD-NHL) [22]. This method initializes the weight matrix with random values and processes it using historical data. Given enough data, this algorithm brings better results that general NHL.

B. Evolutionary Training Algorithms

Evolutionary algorithms, applied in FCM training, make for a data-driven approach: they take use input data and make the FCM adapt to it, minimizing an error function that calculates the difference between the ideal and the computed outputs [23]. They are optimization algorithms, and like all optimization algorithms they are computationally expensive, especially when the solution space is multidimensional, ie when the weight matrix has numerous rows and columns. They're usually applied to compute the optimal weight matrix, but in some cases may also be used to specify the concepts and the relationships among them, and construct the entire FCM from scratch.

Initially, Koulouriotis et al [24] proposed a genetic approach that specifies the FCM structure from data. It uses input-output pairs and computes the structure that will yield the desired outputs, given the corresponding inputs. In order to be effective, this method requires a substantial amount of data containing information on how the system transitions from one state to another.

Parsopoulos [25] applied Particle Swarm Optimization (PSO) for FCM training. His approach uses historical data and converges towards a desired state. PSO uses a swarm of virtual particles that explore the search space. In this FCM training implementation, constraints need to be set in order to keep the FCM structure intact, otherwise it may get modified to the point of losing its meaning, no longer representing the physical system it was suppposed to model.

Petalas et al [26][27] experimented with a memetic approach that combines PSO with deterministic and stochastic local search approaches. Compared to other methods it brought good results.

In [28] PSO was tested in celiac disease (CD) diagnosis on 89 patients and exhibited fast convergence and greater accuracy, in comparison with Bayesian networks, which are usually deployed for this task.

Another proposal from Khan and Chong [29] implements a genetic algorithm that performs backwards engineering and, starting from the desired final state, infers the intitial concept vector.

Mateou et al [30] introduced a genetic optimization algorithm for decision making applications with multiple objectives. This technique computes the weight matrix based on two or more node activation values simultaneously. It improves the previous algorithms, which compute the weights by only taking a single node into account for each iteration, and are challenged when the number of nodes rises.

This is an important problem. As the number of concepts\nodes rises, the usual error function, which computes the sum of deviations from the desired output node values, brings increasingly worse results. Another suggestion to resolve this issue came from Stach et al [31] [32]with an RCGA (Real-Coded Genetic Algorithm) parallel processing algorithm, which aims to train large models, containing dozens of nodes. Using divide-and-conquer tactics it divides the input data into subgroups to assist learning. These algorithms can work without supervision and yield satisfactory results by processing historical data. Regardless, a large number of nodes continues to pose a challenge, as an increased number of variables to be optimized increases rapidly the computational cost. These algorithms may supersede other approaches [33], but there is still great potential for further research and development of new approaches.

Ghazanfari et al [34] deployed Simulated Annealing (SA), which aims for quick computing of the weight matrix, and is superior to GA, in FCMs with large number of nodes. Later on, Alizadeh and Ghazanfari [35] developed a chaotic SA algorithm that proved more efficient than SA, to construct the causality graph, ie the topology of the FCM's edges.

Another method, called DEMATEL (decision-making trial and evaluation laboratory), was introduced by Alizadeh [36]. Its aim was the simplification of the FCM structure by clustering nodes according to the causal relations among them.

Baykasoglu et al [37] trained FCM using the EGDA (extended great deluge algorithm). It's similar to SA and in some cases more preferable, as it takes fewer parameters and has configurable tolerance in suboptimal solutions. This property makes it attractive in applications where lower accuracy would be acceptable, in exchange of lower computational cost and faster convergence.

Ahmadi et al [38] proposed the Imperialist Competitive Learning Algorithm (ICLA), which deploys populations that compete with each other, simulating geostrategic games. The algorithm divides the search space into sectors, extracts information from each sector, and adopts a fast training tactic, avoiding local optima. Its strongest point is that its effectiveness is almost independent of the model's size. It converges fast onto the global otimum and performs well, in comparison with other algorithms.

Chen et al [39] proposed to use Ant Colony Optimization (ACO) and applied it in experiments with FCMs with up to 40 nodes. The ACO algorithm discretizes the continuous search space. Although there is no guarantee it will find the optimal solution, it yields competitive results, in comparison with other algorithms (RCGA, NHL and DD-NHL), in a range of cases.

Another implementation of RCGA, with tournament selection, was used in [10]. They applied the algorithm on timeseries data to identify causal relationships in complex biological systems (gene regulatory networks) with up to 300 nodes. Experimental comparisons with other evolutionary methods (PSO, DE, και AC) show that in highly complex systems it yields, in general, superior results.

Napoles [40] trained FCMs using Random Sampling in Variable Neighborhoods PSO (PSO-RSVN), which surpasses other PSO algorithms [41] and improves on the exploration and exploitation functions during search. It deconstructs the particle swarm from a choice of samples distributed over the search space, each time the seaarch gets stuck or converges prematurely. In [42] the same algorithm was used to customize the sigmoid activation function of each node, resulting in a system that exhibited greater stability.

A different combination of PSO and SSF (Stability based on Sigmoid Functions) was proposed in [43]. This approach seems to improve, in patern recognition applications, the convergence of FCMs that use sigmoid activation functions. The SSF algorithm doesn't depend on PSO, but can be combined with any optimization algorithm.

In [44] Napoles et al developed further the sigmoid customization methodology, where the sigmoid activation functions get optimized without any modifications on the weight matrix. They showed that securing satisfactory accuracy of an FCM is a relatively easy task, as differing output may yield qualitatively similar decisions\classifications.

Later [45], Napoles et al modified the error function and introduced four conditions that specify the accuracy-convergence trade-off. During simulations it was proved that constructing a stable and accurate FCM without modifying the weight matrix isn't always feasible. In the same study, an ensemble strategy for FCMs with 100 nodes and 10.000 weights was tested, with good performance in experiments.

Acampora et al [46] applied the Competent Memetic Algorithm (CMA) and concluded that, compared with other algorithms, CMA is more efficient (requires fewer calculations to converge) and more effective (converges on better solutions than other algorithms). It takes as input the number of nodes and the data from which to learn. The data is represented as a sequence of state vectors. The algorithm constructs a candidate FCM by minimizing the error between the input data and the state vector sequence that results from 'running' the canditate FCM with the same initial state as the input data. It surpasses, statistically, other evolutionary computing techniques, in comparison with GA and PSO, in FCMs with more that 40 nodes.

Kannappan and Papageorgiou [47] presented FCM training with Artificial Immune Recognition System (AIRS) algorithm for pattern recognition. Parameters are set by the user to create a customized model, finetuned for specific problems and requirements. The algorithm showed satisfactory stability and accuracy in a range of parameters, in experiments with a variety of datasets.

Another evolutionary computing approach, the migration algorithms, were utilized by Vascak [48] for the task of training FCMs. In particular, the Self-Organizing Migration Algorithm (SOMA) was used. In SOMA a population explores the search space kai compares the potential solutions it comes across. The population members communicate and interact with each other during search, new populations may be generated, previous populations may be deleted, and the system self-organizes, depending on the progress of the search.

Yesil and Dodurka [49] proposed the Big Bang-Big Crunch (BB-BC) algorithm. It's suggested for decision making scenarios, and finds the optimal initial state that will lead to the desired outcome. Its implementation goes through two phases. First come the Big Bang phase where potential solutions are randomly distributed on the search space, and then follows the Big Crunch phase where the center of mass of the population is computed, oriented towards the direction where the error function is minimized. The procedure is repeated, with each Big Bang phase distributing the populations nearer to the current center of mass. Experiments showed that the algorithm converges quickly and has low computational cost.

Froelich and Papageorgiou [50] introduced a method for univariate time-series prediction FCMs. The method aims to optimize the fuzzification function used for the FCM's concepts. Instead of following the usual practice of transforming all nodes uniformly, this method uses a different, customized fuzzification function for each individual node. This approach differs from other evolutionary computing methods in three respects: (1) it transforms the concepts individually rather than using the same logistic function for all nodes, (2) it optimizes the the fuzzification function rather than keeping it unchanged, and (3) it uses a modified formula to calculate the prediction error.

In 2012, Ma et al experimented with genetic cross-mutation algorithm to train ensemble FCMs for classification tasks [51]. The algorithm, however, exhibited low efficiency and robustness, as well as big latency in the cross-over and mutation functions, leading to slow convergence. In 2015 [52] the researchers published a new method for ensemble models where the FCM uses a learning algorithm, called FCMCM_QC (FCM Cross-Mutation_Quantum Computing), inspired from quantum computing. The algorithm utilizes the concept of the ECQ-bit, where a bit can hold a value of either 0, or 1, or a superposition of both. As the algorithm progresses, the ECQ-bit chromosome converges towards either 0 or 1. Each candidate FCM is assigned an Ensemble Classifier Quantum-bit that converges towards 0 or 1 and specifies whether the candidate FCM will be chosen. FCMs are chosen according to their performance, and the best ones are combined into an ensemble model.

Poczeta et al [53] applied the Structure Optimization Genetic Algorithm (SOGA) so as to construct automatically and optimize an FCM by using multivariate time_series data. The error function contains a penalty coefficient that discourages the model from increasing in complexity, in which case, complexity is defined as the number of nodes and the number of edges. Experimental results showed that this approach achieves tangible reduction of the model's complexity, with the penalty coefficient acting as a threshold that allows only the most significant concepts and interrelations to pass through and participate in the construction of the FCM.

In [54] a new evolutionary multi-objective algorithm was presented. The algorithm picks the most significant nodes according to the direct and indirect influence they exert on the system as a whole. The main difference with other evolutionary algorithms is that while the usual optimization tactic is to minimize an error function, this method also strive to minimize the number of edges in the FCM, as well as to maximize the interactions among concepts. In other words, this algorithm deletes the least significant edges and reenforces the most significant ones in a way that makes up for deleted ones. This variation was proposed because when attention is given exclusively on the error function can result, in many cases, either in modifying the FCM's structure to the point where it no longer reflect the physical system it was supposed to model, or in increasing the systems complexity to extreme levels. Comparative experiments showed that while this method doesn't yield greater accuracy or convergence than others, it does achieve more faithful representation of the physical system.

Wu and Liu conducted research on the problem of FCM training with low volumes of data, which contain noise [55]. They developed the LASSOFCM algorithm that uses the least absolute shrinkage and selection operator (lasso). With this technique the FCM training task is decomposed into a number of sparse signal reconstruction problems, as most weight matrices of FCMs that accurately and realistically capture a system are highly sparse. The method was applied on synthetic data of varying sizes and densities, and yielded good results in FCM learning from time-series with or without noise. It was also applied in gene regulatory networks using the benchmark datasets DREAM3 and DREAM4, with equally encouraging results.

In [56] there is proposed a new evolutionary method that combines the Elite Genetic Algorithm (EGA) and the Individually Directional Evolutionary Algorithm (IDEA), to pick out the most significant concepts and to specify the edge weights, based on the available data. To achieve the model's simplification it uses metrics from Graph Theory, specifically:

the significance of each node (its activation value) and the total influence it exerts (the sum of the weights emanating from it). To train the algorithm the output nodes need to be manually chosen, so that the candidate FCMs will be evaluated according to the error that will be computed for these output nodes. The experimens exhibited reduction in the model's complexity, with acceptable deviation from the desired outputs (activation values of the chosen output\decision nodes).

In 2016, Chi and Liu proposed a multi-objective evolutionary algorithm (MOEA-FCM) to construct FCM from historical data [57]. Other than minimizing the error function, the algorithm also tried to keep the model's complexity in acceptable levels. An unexpected challenge, though, emerged. FCMs with identical degree of complexity were being generated, but their structures were completely different from each other. Additionally, the algorithm required presetting a parameter to guide the search, and the process was afflicted by circularity: in order to find the end solution the parameter should be configured, but to configure it correctly the end solution should be known in advance.

Efforts to alleviate these problems culminated in a new algorithm, proposed in 2019 [58]. The new methodology combines ensemble FCMs and a multi-objective evolutionary algorithm (MOEA) to reconstruct gene regulatory networks. The algorithm starts by constructing, based on historical data, a number of FCMs and then chooses those with the lowest error on the training data. Finally, the chosen FCMs are combined into a final, ensemble FCM. Other than minizing the error function, the algorithm also tries to optimize the FCM's internal structure. Since FCMs share many commonalities with neural networks, and many techniques from neural networks can be implemented in FCMs, the researchers built on previous research from Abass [59], who developed a method of constructing ensemble neural networks. In experimental tests the algorithm exhibited efficiency and effectiveness.

Also from Chi and Liu, another implementation for gene regulatory networks reconstruction [60] uses a combination of memetic algorithms and neural networks, called MANN (Memetic Algorithm Neural Network) FCM-GRN. The memetic algorithm detects the existence of connections among the FCM nodes and the neural network determines the strength of these connections. Experimental comparisons showed that this methodology surpasses the ACO, NHL, and RCGA algorithms.

Most of the FCM learning algorithms that have been proposed focus on small-scale FCMs and generate FCMs that are denser than the real-world modeled system. As the system's complexity increases and the weight matrix expands, the learning becomes exponentially harder. On this direction, Zou and Liu proposed the MIMA-FCM (mutual information-based memetic algorithm FCM) algorithm [61], which addresses large-scale FCMs and is implemented in two phases. During the first phase the algorithm detects whether there are causal interactions among the system's nodes\concepts. Keeping the system's density on low levels, the search space is drastically reduced. In the second phase, after the search space has been rendered as accesible as possible, the memetic algorithm optimizes the edge weight values. This method was tested on gene regulatory networks and exhibited encouraging results on large-scale FCMs.

A proposal on FCMs for classification [62] adds to the learning algorithm a further action. After the weight matrix has been generated and optimized, a further optimization is done on the decision thresholds of the FCM's outputs. This addition, as the experiments show, brings higher performance on classification problems.

Hatwagner et al [63] combined a genetic optimization algorithm with clustering techniques to unify similar concepts into groups. The aim was the simplifying of the emerging FCM for better interpretability, as oversized FCM are hard to be understood by humans. By grouping the semantically similar concepts, which exert qualitatively and quantitatively similar influences on the system, the simplification of the FCM that can be achieved is substantial. Oversimplification, though, leads to loss of useful information and to degradation of performance, hence the acceptable trade-off between simplicity\interpretability and lost information\lower performance needs to be determined by the user.

Poczeta et al [64] focused on how FCMs that are consructed from historical data are denser than those made by humans. In this context, 'denser' means the number of edges among nodes is much greater, in proportion, than the number of nodes. The researchers developed a genetic algorithm that optimizes the system's performance while accounting for its density. In this way they achieved an algorithm which, by retaining only the most significant edges among nodes, constructs automatically an FCM similar to one humans would make.

In 2018, Wu and Pallath patented an FCM learning algorithm based on the artificial immune system (AIS) method [65]. Initially the algorithm detects the concepts\nodes, specifies the value of each concept, for each moment in the sequence, and generates an initial FCM structure and an initial weight matrix. Then it generates a sequence of potential weight matrices, computing the system's response for each one, evaluates the results and, deploying the AIS algorithm, converges on the optimal weight matrix.

Salmeron et al [66] proposed the Modified Asexual Reproduction Optimisation (MARO) algorithm. In contrast with the various population-based αλγόριθμους, MARO examines only one candidate solution, and computes a single error function, each time. This brings a drastic reduction on execution speed. The algorithm avoids local optima by utilizing the SA methodology to explore a wider range of possible solutions, and due to the automated path choice, it doesn't require calibrating hyperparameters. Experimental results revealed that the algorithm is fast and achieves greater accuracy than other techniques.

Yang and Liu [67] emphasize that training a single FCM isn't the ideal solution, when it comes to complex and dynamic systems, because the FCM may just reflect a local optimum. They suggest training multiple FCMs, by combining the multi-agent genetic algorithm (MAGA), which is a numerical optimization algorithm, with niching techniques. With this combination they developed the NMM-FCM algorithm (niching-based multi-modal multi-agent genetic algorithm), where multiple FCMs are trained, and the optimal one is handpicked. Experiments showed that the algorithm achieves high accuracy in gene regulatory network reconstruction tasks.

B. Hybrid Training Algorithms

Hybrid FCM learning techniques combine Hebbian learning with evolutionary methods. They start with expert knowledge and Hebbian learning, and after a preliminary FCM is constructed, they deploy evolutionary computing for further modification and improvement upon the weight matrix. In [68] DE and NHL were combined. Zhu and Zhang [69] combined RCGA kai NHL. NHL and Extended Great Deluge Algorithm (EGDA) were tested in [70], where NHL computed a weight matrix approximating the desired outcome, then finetuned it with the EGDA, minimizing the error function. Natarajan et al [71] combined DDNHL and GA, developing a method called FCM-DDNHL-GA, and showed that it surpasses other approaches on classification tasks.

# IV. Discussion

FCM learning aims, as a general rule, to specify the weight matrix, which maps the causal relationships among a system's concepts, and quantifies (in fuzzy set range) these relationships. Each FCM learning algorithm provides its own advantages and faces its own challenges. Choosing the appropriate algorithm depends on the task at hand, the type and degree of complexity of the individual real-world system that needs to be modeled, our time and computational power constraints, and the appropriate trade-off between accuracy and generalization.

Hebbian learning is used when the FCM has already been constructed, based on expert knowledge, and afterwards through training the FCM we process and fine-tune the weight matrix. The FCM retains its structure to a great extent, thus the causal relationships among the system's variables, as described by the experts, remain in place. With Hebbian learning fine-tuning of the weight matrix occurs, but no qualitative change is effected. Thus the FCM retains its meaning, its interpretability and its reliability in decision making scenarios. Basic disadvantage of this method is its low generalization capability, especially when the FCM edge weights are suboptimal [72].

Recommended uses of Hebbian learning algorithms are control tasks, where the constraints and the desired equilibrium point are clearly known. Evolutionary learning algorithms are recommended for pattern classification and forecasting tasks, provided there are enough data to train the FCM adequately [72].

During the last years there has been a trend towards genetic and evolutionary algorithms for FCM training, and research in this field has blossomed substantially in contrast with Hebbian learning and the various hybrids, which have remained relatively static.

Why Evolutionary Algorithms

Main reason for the supremacy of genetic and evolutionary algorithms for FCM training is to achieve independence from human experts. Extracting knowledge from specialists is time-consuming and often unreliable. [73]. Their knowledge on their particular field is mostly subjective and intuitive and, even if we could find and motivate them to cooperate for the creation of an FCM, which isn't always feasible, there are no guarantees that the extracted information will be useful, or even correct. Different experts will give instructions, on a particular field, that will result in totally different FCMs, which is an interesting example of human subjectivity. As the system that needs to be modeled rises in complexity, human opinions grow increasingly unreliable. A lot has been written on this subject [74][75][76][77][78][79], many proposals have been made, many studies for reliable knowledge extraction from experts have been conducted, many approaches have been tried and tested, and the end result of all these is the newly emerging trend towards evolutionary algorithms that can learn from historical data without human intervention.

Even if constructing FCMs from human experts was more reliable, efforts to automate the task would still be extremely worthwhile. In that case the task turns into a combinatorial optimization problem and real-valued coefficient approximation, in other words, finding the values of the FCM's weights, as well as the structure of their combination into a matrix.

This is exactly the type of optimization in which genetic andevolutionary algorithms excel. "Unlike some approaches," Coley writes [80], "their promise has rarely been over-sold and they are being used to solve practical problems on a daily basis."

In optimization problems the number of possible solutions rises exponentially with the number of variables. Exhaustive search soon becomes impossible and simple methodologies from differential calculus depend entirely upon of the search's starting point: if a local optimum is interpolated between the starting point and the global optimum, the search fails. As the task at hand gets interesting (ie the number of parameters increases), approaches like these become irrelevant. This is where random search techniques and simulated annealing have been proposed [81]. Genetic algorithms start from random search and direct it using intelligent meta-heuristics inspired from evolutionary theory and other scientific observations of the natural world.

Evolutionary computing cannot, theoretically, guarantee it will find the optimal, or even an acceptable, solution, but in practice it's often quite successful and is widely used in industry and applied science [82]. In multi-variate optimization problems, where the variables affect the output non-linearly, evolutionary algorithms are the established solution, and automated FCM learning falls under this problem category.

Choice of Algorithm

It is difficult to choose the best algorithm for general FCM training because most publications have focused on relatively simple problems and different applications [10]. Addressing a particular problem, in the case of using evolutionary algorithms for learning FCMs, means customizing the error function. For example, in [53] a penalty coefficient is assigned to every node and non-zero weight, keeping the system's complexity low. For pattern recognition tasks the error function is only applied on the output nodes [83], and generally, for decision making applications, the error function focuses mainly on the nodes of interest, thus providing more accuracy on these nodes. For some applications we may wish to contain the potential outputs of certain nodes within a specific range, a constrained which could be encoded on the error function.

The standard error function calculates the distance between desired and calculated outputs, and the goal of training is to minimize this distance. Many error-functions are non-differentiable, making gradient-based learning methods inaccessible, and leaving evolutionary computing as the only option. In [23] a variety of error functions used in FCM learning are discussed.

Challenges and Solutions

As the search space expands, the computational cost of evolutionary algorithms rises. This cost could be minimized by optimizing the algorithms at computer code level. A big advantage of evolutionary computing is its easy parallelization. With the late availability of GPUs for fast computations and vector-matrix multiplications, the computational cost could be drastically reduced.

Another challenge is the possibility of missing the global optimum and converging to a local one. Owing to the very nature of evolutionary computing, this problem will be solved not through theoretical analysis but empirically, with experimental tests and comparisons of various approaches. The cooperation between genetic algorithms and Simulated Annealing would help make progress whenever this problem arises. Generally, and for each unique circumstance, the optimal proportion of exploration and exploitation must be determined through experimentation, with the search step being configured to meet the requirements.

Further research is required for constructing more complex FCMs. As the number of nodes and non-zero weights rises, the model's performance deteriorates. Small deviations in the weight values result in large deviations in the final output. To improve accuracy, local search functionalities could be embedded in the learning algorithms, keeping the computational cost into account. According to [46], some questions that need to be answered for each specific application are: (1) how often will local search need to be conducted in each iteration\generation, (2) when should it be done, (3) which members of the population should be used, (4) what computational resources will be allocated for each local search, and (5) which method will be used.

Generally, searching close to a population member refines that member and gives greater precision to its value, and searching further away from it helps to escape local optima and to effectively explore the search space in its entirety. The right proportion among these two types of search, the right time and the right occasion for each, is determined empirically and is suited to each particular application. This personalization seems to be the key to effective FCM learning through evolutionary computing techniques.

## V. Conclusion

FCMs provide capabilities that other machine learning models lack, and especially: (1) their interpretability, and (2) their ability to capture and represent the causal relations and hierarchical structure of a system. Improving the model and its techniques may open new horizons and bring potential and opportunities that other machine learning models don't have access to. Research towards the improvement and establishment of FCMs would be a highly fruitful endeavor.

According to the research on the field so far, the main challenges seem to be the difficulty in building FCM with large number of nodes and edges, generalization, and the speed of convergence towards high-quality solutions. Future research would be well directed in these issues.

With the advent of big data, the internet, the new technologies, and the modern trend towards digitization of an increasing volume of information, we have now access to an unprecedented amount of data from a wide range of systems and knowledge fields [84]. A huge amount of this information and data could be used for constructing FCMs that model real-world systems from all aspects of life. Extremely complex systems, which could never be fully grasped by humans, can now be, through the use of evolutionary learning algorithms, mapped as FCMs. A fruitful direction research on FCMs could, and should, take, is their scaling up towards being able to absorb increasing volumes of data and reliably representing increasingly complex systems.